# Cognitive Mapping of Learning by Thinking Mechanisms in Large Language Models

**Dong Han[1,2], Jin Zhang[3*], and Kunyue Yang[1]**

[1]School of Information Science and Technology, Northeast Normal University, Changchun, Jilin, China
[2]Office of Information Management and Planning, Northeast Normal University, Changchun, China
[3]American Scholars Press, GA, USA
[*]Corresponding author: jinusa2000@yahoo.com
Dong Han: handong@nenu.edu.cn. Kunyue Yang: kyyang888@nenu.edu.cn

**Abstract**

Large language models (LLMs) increasingly exhibit capabilities that resemble human Learning by Thinking (LbT), the acquisition of new insight through internal processes such as explanation and reasoning, without new external input. A systematic account of the cognitive mechanisms behind these capabilities, and of their limits, remains underdeveloped. This paper proposes a Cognitive LbT Framework that maps four mechanisms (Reflective Reasoning, Synthetic Memory Formation, Internal Simulation, and Conceptual Self-Organization) onto computational processes in LLMs. Drawing on representative methods, including chain-of-thought prompting, self-consistency, self-reflection, Tree-of-Thoughts search, Self-Instruct, and self-taught reasoning, the analysis argues that selected methods can functionally approximate aspects of these mechanisms and can improve performance on some tasks without additional labeled examples, although this does not always constitute durable learning in model parameters. For example, Self-Instruct generated approximately 52,000 instructions from 175 seed tasks and reported a 33% absolute improvement on Super-NaturalInstructions (Y. Wang et al., 2023). The same mechanisms also produce characteristic failures, including hallucination diffusion, model collapse, concept drift, and high computational cost. These failures are not isolated technical defects but follow from four structural asymmetries between LLMs and human cognition: the identity of generator and monitor, closed learning loops without external grounding, symbolic rather than sensorimotor simulation, and the cost of inference-time computation. The framework yields testable propositions about when LbT in LLMs should succeed or fail and offers a structured basis for designing more reliable autonomous learning systems.

## 1. Introduction

Large language models (LLMs) have progressed from text generators to systems capable of multi-step reasoning, self-critique, and the generation of new training material (Brown et al., 2020; Wei et al., 2022). This progress raises a question that is central to both artificial intelligence and cognitive science: can LLMs, like humans, gain new insight through internal processes alone, without new external input? In cognitive science, this capacity is studied as Learning by Thinking (LbT) (Lombrozo, 2019, 2024). Learners improve their understanding by explaining material to themselves, and scientists refine theories through thought experiments, in both cases without new observations (Chi et al., 1989; Nersessian, 2008). LbT thus offers a way to ask what LLMs gain when they "think" and what they cannot gain in that way.

The empirical record is mixed. Chain-of-thought prompting elicited multi-step reasoning abilities (Wei et al., 2022), and methods such as Reflexion (Shinn et al., 2023), STaR (Zelikman et al., 2022), and related self-training approaches (Singh et al., 2024) produced performance gains without additional labeled data. Other studies show that, without external feedback, self-correction can fail to remove errors and can even degrade performance (Huang et al., 2024), and that recursive training on self-generated data degrades models over successive generations (Shumailov et al., 2024). The field has largely debated these results in ontological terms, asking whether LLMs "really" reason (Bender & Koller, 2020; Mitchell & Krakauer, 2023). Less attention has been paid to a mechanistic question that existing evidence can address: which components of human LbT have functional counterparts in LLMs, how do these components interact, and why do they fail where they fail?

This paper addresses that question by proposing a Cognitive LbT Framework that maps four mechanisms (Reflective Reasoning, Synthetic Memory Formation, Internal Simulation, and Conceptual Self-Organization) onto LLM computational processes. The paper makes four contributions:

- **Theoretical integration.** It grounds the four mechanisms in the cognitive science of LbT and maps each onto its computational implementations, providing a unified taxonomy for methods that are usually studied in isolation (Section 3).
- **Operational specification.** For each mechanism, it specifies implementations, process-level evaluation indicators, and characteristic failure modes, and it describes how the mechanisms interact through recurrent feedback (Sections 3.1 and 3.6).
- **Explanatory account of failure.** It argues that the main failure modes of LLM-based LbT follow from four structural asymmetries with human cognition, and it derives testable propositions from this account (Section 4).
- **Design implications.** It derives design principles (integrated representation, recurrent feedback with external anchoring, and diversity maintenance) and research directions for more reliable autonomous learning systems (Section 5).

As a theoretical paper, it reports no new experiments. It integrates existing findings into a framework intended to organize, explain, and generate predictions about LbT in LLMs. Section 2 reviews the human cognitive foundations of LbT and the debate over LLM cognition. Section 3 presents the framework and maps each mechanism. Section 4 develops the structural-asymmetry account of failure. Section 5 discusses achievements, implications for cognitive science, design principles, research directions, and limitations, and Section 6 concludes.

# 2. Theoretical Background

## 2.1 Human Foundations of Learning by Thinking

Research on LbT draws on work in cognitive science and machine learning on explanation-based learning and conceptual change (Carey, 2009; Chi et al., 1989; DeJong & Mooney, 1986). In contrast with behaviorist accounts that emphasize external reinforcement and association (Skinner, 1938; Thorndike, 1911), LbT concerns "offline" cognition: learning that proceeds by reorganizing knowledge the learner already has. Learning by Thinking denotes the acquisition of new insight through internal processes such as explanation, mental simulation, analogical comparison, and reasoning, without new external input (Lombrozo, 2019, 2024). Through these processes learners integrate fragmented knowledge, draw inferences beyond immediate experience, and construct abstract schemas, so that cognitive growth results from restructuring existing knowledge rather than from accumulating new input.

Evidence for the value of LbT comes from research on self-explanation. Chi et al. (1989) found that students who generated more self-explanations while studying worked examples solved subsequent problems more successfully than students who generated fewer. Later work describes how explaining mechanisms can help learners identify gaps, integrate fragmented information, and revise mental models (Chi, 2000). Because the size of these effects depends on the task and the measure, no general percentage advantage is asserted here.

The cognitive resources on which LbT draws develop early and continue to develop across the lifespan. Children as young as two to four years construct causal maps from patterns of evidence and can use them to predict the effects of interventions (Gopnik et al., 2004). In adulthood, LbT appears as sophisticated mental simulation. Scientists use thought experiments, such as Einstein's imagined pursuit of a light beam, to probe the logical limits of a theory without physical observation (Nersessian, 2008), and people judge outcomes by mentally simulating alternative courses of events (Kahneman & Tversky, 1982). Simulating counterfactual physical processes ("What if the ball had been thrown at a different angle?") allows learners to internalize causal structure and generalize to new situations.

At the neural level, self-generated thought of the kind involved in LbT recruits the default mode network together with the frontoparietal control network (Andrews-Hanna et al., 2014). The default mode network supports internally directed cognition and episodic simulation, and it overlaps with the networks used to remember the past and imagine the future (Schacter et al., 2007). Prefrontal control systems maintain information in working memory and monitor logical consistency. Their interplay balances divergent imagination with goal-directed evaluation and gives human LbT an internal check on plausibility, which becomes important in Section 4.

Human LbT is thus multilayered: it ranges from everyday self-explanation to expert thought experiments, from concrete causal inference to abstract theoretical synthesis, and from simulation-oriented to control-oriented neural systems. The common denominator is cognitive growth through the internal reorganization of existing knowledge. A second, easily overlooked feature is that human LbT is continually checked against perception and social interaction (Vygotsky, 1978). Both features are central to the framework developed below.

## 2.2 The Debate Over LLM Cognition

Recent LLM methods show apparent functional parallels to human LbT. Chain-of-thought prompting resembles self-explanation, self-consistency resembles the search for convergent evidence, self-reflection resembles metacognitive monitoring, synthetic-data generation resembles internal knowledge synthesis, and tree search resembles deliberate mental simulation. Table 1 summarizes the representative methods analyzed in this paper. Whether these parallels reflect cognition or sophisticated statistical reproduction remains contested. The debate turns on three questions: the grounding of meaning, the mechanics of error, and the nature of emergent reasoning.

**Table 1 | Representative LLM Methods Analyzed in This Paper and Their Mapping to LbT Mechanisms.**

| Method | Source | Core operation | Reported result | Mechanism |
|---|---|---|---|---|
| Chain-of-thought (few-shot and zero-shot) | Wei et al. (2022); Kojima et al. (2022) | Generates explicit intermediate reasoning steps before the answer | Effects vary by task, model, and prompting setup | Reflective Reasoning |
| Self-Refine | Madaan et al. (2023) | Generates, self-critiques, and revises iteratively | ~20% absolute average improvement across 7 tasks; code-generation gains up to 13 points with a code model (Codex) in tested setups | Reflective Reasoning |
| Reflexion | Shinn et al. (2023) | Verbal self-reflection stored across trials | 91% pass@1 on HumanEval in the reported setup | Reflective Reasoning |
| Self-Instruct | Y. Wang et al. (2023) | Generates new instructions from seed tasks for fine-tuning | 175 → ~52,000 generated instructions; +33% absolute on Super-NaturalInstructions | Synthetic Memory Formation |
| STaR | Zelikman et al. (2022) | Fine-tunes on self-generated rationales that reach correct answers | Iterative accuracy gains on arithmetic and commonsense reasoning | Synthetic Memory Formation |
| Tree-of-Thoughts | Yao et al. (2023) | Branches, evaluates, and searches over intermediate thoughts | Game of 24: 4% → 74% | Internal Simulation |
| Self-consistency | X. Wang et al. (2023) | Samples many reasoning paths and takes the majority answer | Improved GSM8K in tested settings; magnitude varies by model | Internal Simulation |
| Representation learning | Mikolov et al. (2013); Zelikman et al. (2022) | Relational structure and abstraction in learned representations | Vector-offset analogies; transfer to held-out problems | Conceptual Self-Organization |
| Voyager | G. Wang et al. (2023) | Curriculum, skill library, and environment-feedback loop | Lifelong skill acquisition and composition | All four (integration) |

*Note.* Reported results are task- and model-specific; consult the cited primary studies for the relevant models, benchmarks, and evaluation settings.

### *Semantic Grounding and Functional Competence*

Bender and Koller (2020) argue that a system trained only on linguistic form, without physical or social interaction, can at best manipulate symbols and cannot achieve genuine semantic understanding. Mahowald et al. (2024) refine this point by distinguishing formal linguistic competence, at which LLMs often approach human performance, from functional competence, which requires modeling real-world causality and social intention. From this perspective, LLMs lack the grounded world model on which human LbT relies.

### *Epistemic Fragility*

Skeptics also question the reliability of LLM self-improvement. Human reflection is often self-correcting, but LLM self-correction without external feedback frequently fails to improve, and can degrade, reasoning performance (Huang et al., 2024). Internal reasoning can rationalize and amplify an initial error. This vulnerability is consistent with the "stochastic parrot" critique (Bender et al., 2021) and with broader arguments that deep learning lacks robust abstraction (Marcus, 2018, 2020): what appears to be thinking may be a high-dimensional recombination of training patterns, which raises the question of whether such systems can generate new knowledge at all (Mitchell & Krakauer, 2023).

#### *Emergent Cognitive Structure*

Other evidence suggests that LLMs capture more than surface correlations. Neural networks can achieve human-like systematic compositional generalization when trained appropriately (Lake & Baroni, 2023), and language models can be understood as modeling the beliefs and intentions of the agents who produce text (Andreas, 2022). Such findings suggest that training on large text corpora can lead models to internalize some structural properties of the world even without embodiment.

#### *Toward a Task-Centric Synthesis*

Shanahan (2024) recommends replacing debates about the "essence" of LLM cognition with careful analysis of what these systems do, under which conditions, and where they fail. This paper adopts that stance. It does not ask whether LLMs think in the human sense; it asks how their internal processing reorganizes information, which functional correspondences with human LbT hold, and which structural differences explain the observed failures.

## 2.3 Research Gaps and Positioning

Three gaps motivate the framework.

- **Fragmentation.** Cognitive science has a well-developed account of human LbT (Section 2.1), and AI research has produced many separate techniques (Table 1), but the two literatures are rarely integrated. A notable exception is Lombrozo (2024), who brings natural and artificial minds into a single discussion of LbT; the integration has not yet been carried to the level of specific LLM mechanisms. As a result, little is known about how techniques such as chain-of-thought prompting and self-reflection interact or which human mechanisms they correspond to.
- **Lack of mechanistic mapping.** Cognitive terms such as "reflection," "memory," and "simulation" are widely used in AI research, but their computational counterparts are rarely defined. Debates often stall on whether LLMs "truly understand" without analyzing the functional equivalences and architectural differences between biological and artificial systems.
- **Absence of an explanatory account of failure.** Related failure modes, such as the spread of errors during self-correction (Huang et al., 2024) and model collapse (Shumailov et al., 2024), are documented. Yet there is no theoretical explanation of why LLM self-correction sometimes amplifies errors while human reflection usually reduces them, or why LLMs succeed at complex abstract reasoning yet fail at simpler tasks that require real-world grounding.

The Cognitive LbT Framework addresses these gaps and extends Lombrozo's (2024) comparison in three respects. It shifts the question from whether LLMs think to how they process and reorganize internal information, mapping LbT onto specific LLM mechanisms, each with implementations, evaluation indicators, and failure modes; it explains the failures of LLM-based LbT through a small number of structural asymmetries with human cognition; and it derives testable propositions from that explanation.

# 3. The Cognitive LbT Framework

## 3.1 Derivation and Guiding Principles

The framework builds a functional bridge between human cognitive processes and LLM operations. To organize the four mechanisms, this paper proposes an analytical distinction, drawn from the human LbT literature reviewed in Section 2.1, between two kinds of process. Generative processes produce new candidate knowledge by operating on existing representations; in human LbT they include explanation and reasoning, as well as simulation and counterfactual thinking. Consolidating processes turn the products of thinking into more durable knowledge; they include constructive memory, in which imagined and recalled events share representational resources (Addis et al., 2007; Bartlett, 1932), and conceptual change, the reorganization of conceptual structure that LbT can produce (Carey, 2009; Nersessian, 2008).

On this basis, the framework treats Reflective Reasoning (grounded in explanation and reasoning) and Internal Simulation (grounded in simulation and counterfactual thinking) as generative mechanisms, and Synthetic Memory Formation (grounded in constructive memory) and Conceptual Self-Organization (grounded in conceptual change) as consolidating mechanisms. The distinction is a proposal of the present framework, not an established classification in cognitive science, and it identifies the primary function of each mechanism, not an exclusive one: reflection can also stabilize knowledge, and consolidation can expose gaps that prompt further reasoning. Table 2 summarizes the resulting mapping.

The construction of the framework follows three principles.

- **Functional equivalence at the computational level.** Following Marr's (1982) levels of analysis, the framework compares mechanisms by their computational goals rather than by their physical implementation. The criterion is not biological replication but whether the artificial system fulfills the same cognitive function, for example error detection and correction, as its human counterpart.
- **Emergence.** LbT-like capabilities often appear as properties of scale rather than of explicit programming. Chain-of-thought reasoning, for example, has been reported to yield gains mainly in models above roughly 100 billion parameters (Wei et al., 2022). Whether such abilities emerge abruptly or only appear abrupt because of the metrics used to measure them is debated (Schaeffer et al., 2023), so the framework treats scale as an enabling condition and not as a sharp threshold.
- **Operationalizability.** To avoid purely metaphorical description, each mechanism must be specified by concrete implementations, process-level evaluation indicators, and identifiable failure modes. This makes the framework usable both as a theoretical model and as a diagnostic tool.

**Table 2 | Derivation and Mapping of the Four LbT Mechanisms.**

| Mechanism | Human LbT basis | LLM implementations | Process-level evaluation indicators | Characteristic failure modes |
|---|---|---|---|---|
| Reflective Reasoning (generative) | Self-explanation, reasoning, metacognitive monitoring (Chi, 2000; Flavell, 1979) | Chain-of-thought; Self-Refine; Reflexion | Step-level correctness; faithfulness of reasoning to answer; calibration of self-critique | Hallucination diffusion; self-reinforcing error spirals; performative reasoning |
| Synthetic Memory Formation (consolidating) | Constructive memory; imagined events share substrates with recalled ones (Addis et al., 2007; Bartlett, 1932) | Self-Instruct; STaR; self-training | Accuracy of retained data; diversity across iterations; retention of prior abilities | Model collapse; hallucination consolidation |

| Mechanism | Human LbT basis | LLM implementations | Process-level evaluation indicators | Characteristic failure modes |
|---|---|---|---|---|
| Internal Simulation (generative) | Mental simulation and counterfactual thinking (Johnson-Laird, 2006; Kahneman & Tversky, 1982) | Tree-of-Thoughts; self-consistency | Counterfactual sensitivity; consistency across branches; computation per correct answer | Grounding failures; search explosion; inconsistent branches |
| Conceptual Self-Organization (consolidating) | Conceptual change and abstraction (Carey, 2009; Rosch, 1978) | Representation restructuring through training and self-training | Transfer to held-out problems; stability of concepts across iterations | Catastrophic forgetting; shallow semantics; concept drift |

*Note.* Evaluation indicators are proposed by the present framework; Section 5.4 discusses benchmarks that could operationalize them. Failure-mode terms are defined at the beginning of Section 4.

The four mechanisms do not operate in isolation. Figure 1 shows their logical order and the feedback loops that connect them. Reflective Reasoning performs sequential inference through chains of thought and iterative self-critique. Synthetic Memory Formation distills successful reasoning trajectories into reusable, long-term representations. Internal Simulation generates and evaluates alternative scenarios to predict outcomes in new situations. Conceptual Self-Organization extracts abstract principles from these outputs and reorganizes the model's internal semantic structure, which in turn shapes subsequent reasoning. The order alternates generation and consolidation: each generative step supplies material that the next consolidating step stabilizes. Through this loop the system can, in principle, refine its knowledge without continuous external supervision. Section 4 argues that the qualification "in principle" matters: the loop is reliable only when it is coupled to signals from outside the model.

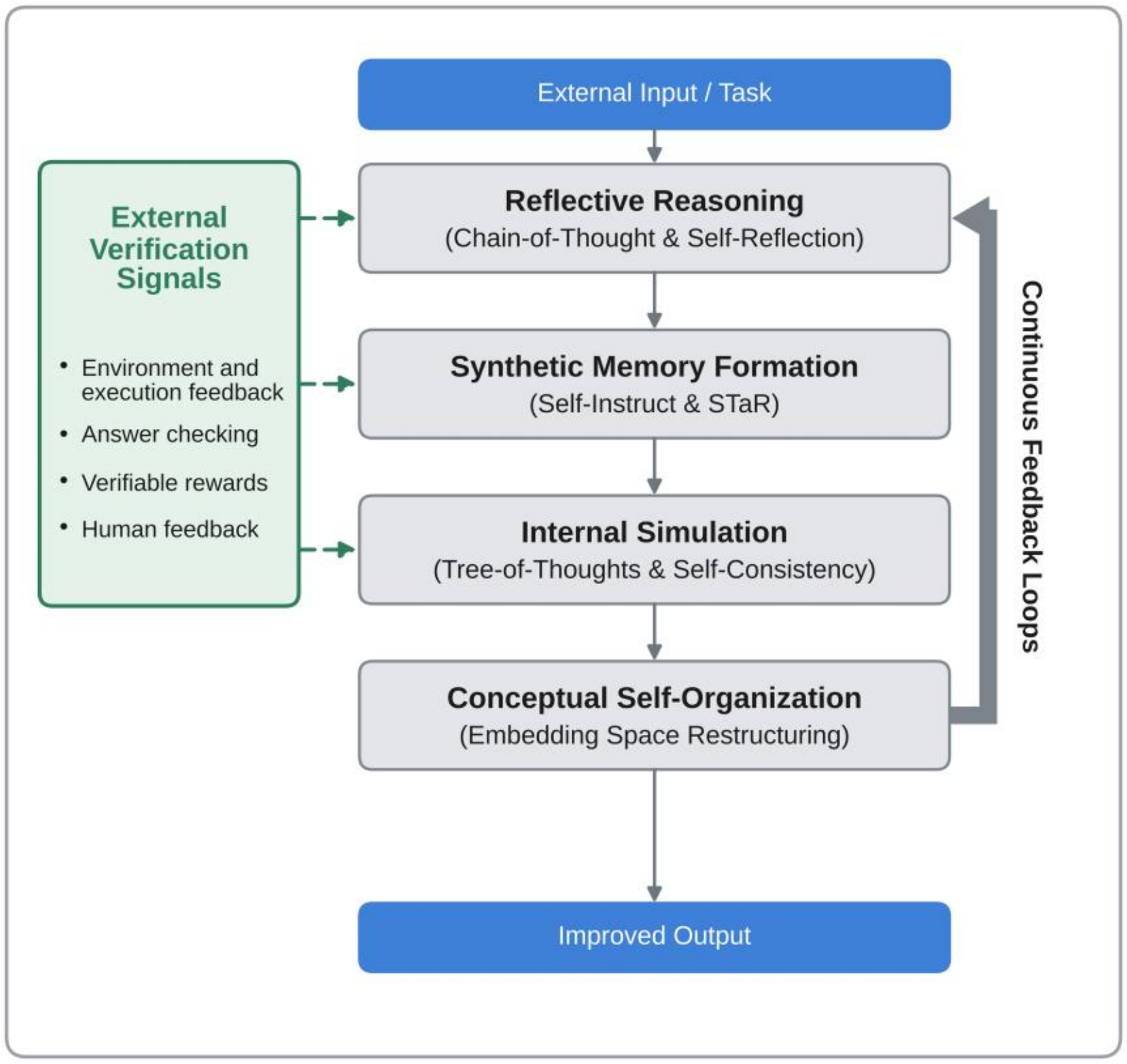

**Figure 1 | Overall Architecture of the Cognitive LbT Framework.** Note. The four mechanisms are shown in their logical order, alternating generation (Reflective Reasoning, Internal Simulation) and consolidation (Synthetic Memory Formation, Conceptual Self-Organization); the side arrow indicates the continuous feedback loops that connect them. The green panel represents external verification signals, and the dashed arrows mark where these signals anchor the loop outside the model (see Section 4.2).

## 3.2 Reflective Reasoning: Self-Explanation and Metacognitive Loops

### *Human Function*

Reflective Reasoning is the capacity to improve internal knowledge through self-generated explanation, critical appraisal, and belief revision (Chi, 2000; Fleming & Dolan, 2012). Reflection deepens comprehension and exposes blind spots, even when the initial explanation is incomplete, because the act of explaining reveals what the learner does not yet understand.

### *LLM Implementation*

In LLMs, Reflective Reasoning is implemented in two complementary ways: linear reasoning trajectories and iterative optimization loops.

**Chain-of-thought as functional self-explanation.** Chain-of-thought prompting (Kojima et al., 2022; Wei et al., 2022) elicits explicit intermediate steps for complex problems. These steps can make parts of a solution easier to inspect and check, although their presence does not guarantee that the stated reasoning faithfully explains the model’s answer. Reported effects vary with model scale, task, prompting setup, and evaluation metric (Wei et al., 2022). One possible functional interpretation is that intermediate steps externalize some information that would otherwise have to be maintained within a single generation (Baddeley, 2000; Sweller et al., 1998).

**Self-reflection as iterative optimization.** Self-reflection loops add a second layer (Madaan et al., 2023; Shinn et al., 2023). The process typically has four stages: initial generation; critique, in which the model identifies possible flaws; revision based on that feedback; and repetition until a stopping condition is met. Across seven tasks, Self-Refine reported an average improvement of approximately 20% absolute in task performance, with task-specific effects varying substantially (Madaan et al., 2023). In a separate study, Reflexion achieved 91% pass@1 on HumanEval in the reported setup (Shinn et al., 2023). These results show that iterative feedback can improve selected outputs; they do not establish a universal advantage over reinforcement learning. The distinctive feature of both methods is the use of natural-language feedback as an intermediate representation in an iterative generation-and-revision loop.

**Task dependence.** The benefits of self-reflection depend on the task, the quality of the feedback, the base model, and whether errors can be checked against an external or verifiable signal. Improvements observed on selected complex tasks should not be treated as a universal effect across benchmarks.

### *Characteristic Failure*

Reflection improves outputs only when the critique stage can detect errors. When the same model generates and evaluates, early errors can be rationalized rather than corrected, and reasoning chains can be performative rather than faithful to the computation that produced the answer (Turpin et al., 2023). Section 4.1 analyzes these failures and explains why they follow from the architecture. Functional correspondence at the level of outputs therefore does not imply mechanistic equivalence: human reflection is regulated by awareness of uncertainty, whereas LLM reflection lacks an independent check on its own premises.

## 3.3 Synthetic Memory Formation: From Inference to Consolidation

### *Human Function*

Human memory actively reconstructs information rather than storing it passively (Bartlett, 1932). Part of this process is the construction of synthetic experience: imagined events that recruit neural substrates overlapping with those of veridical memories (Addis et al., 2007; Schacter et al., 2007). Imagined and recalled experience thus appear to draw on overlapping resources that later reasoning can use.

### *LLM Implementation*

LLMs implement a functional analogue by generating their own training data and consolidating it into their parameters. Self-Instruct (Y. Wang et al., 2023) expanded 175 seed tasks into about 52,000 instructions and reported a 33% absolute improvement on Super-NaturalInstructions. STaR (Zelikman et al., 2022) uses a bootstrap procedure: the model generates chain-of-thought rationales, keeps those that reach correct answers, and is fine-tuned on them iteratively; related self-training methods extend this procedure to larger problem sets (Singh et al., 2024). These methods convert transient reasoning trajectories into persistent knowledge. Reasoning that first required explicit deliberation becomes available as consolidated pattern recognition, which this paper calls "computational intuition."

### *Characteristic Failure*

Synthetic Memory Formation is vulnerable to model collapse: when models are trained recursively on their own outputs, the tails of the original data distribution disappear and output diversity shrinks across generations (Shumailov et al., 2024). Without an external source of truth, errors produced during generation are also written into the weights during fine-tuning, a process this paper calls hallucination consolidation. Human memory construction is continually corrected by environmental and social feedback, whereas a closed synthetic loop lacks that correction and risks “digital autophagia,” the term this paper uses for a system feeding on its own outputs. Section 4.2 analyzes these failures together with concept drift.

***Transition to Internal Simulation***

Consolidated synthetic memories supply templates for further reasoning, but stored patterns do not themselves show whether they hold in new situations. Testing that requires the next mechanism, Internal Simulation.

## 3.4 Internal Simulation: Parallel Exploration and Deliberate Search

***Human Function***

Internal Simulation allows people to run hypothetical scenarios mentally and infer their consequences without acting (Johnson-Laird, 2006; Kahneman & Tversky, 1982). It supports causal learning and strategic planning.

***LLM Implementation***

In LLMs, Internal Simulation is achieved by generating, branching, and evaluating alternative reasoning trajectories.

**Deliberate tree search.** Tree-of-Thoughts (Yao et al., 2023) moves the model from single-pass generation to deliberate search over a space of intermediate “thoughts.” The procedure generates candidate reasoning steps, evaluates the prospects of each path, retains the most promising candidates (for example, by breadth-first or beam-style search), and iterates until a solution is reached. This shift resembles the move from fast, intuitive “System 1” processing to deliberate “System 2” processing (Kahneman, 2011). On the Game of 24, GPT-4 with Tree-of-Thoughts reached a 74% success rate, compared with 4% under chain-of-thought prompting; in a creative-writing task, its passages also received higher coherence ratings than those produced with chain-of-thought prompting (Yao et al., 2023).

**Convergent sampling.** Self-consistency (X. Wang et al., 2023) samples many reasoning paths from the same model (40 in the original experiments) and selects the most frequent answer. This approach has improved accuracy on GSM8K, a benchmark of grade-school mathematics problems (Cobbe et al., 2021), in tested model settings, although the magnitude varies by model and benchmark (X. Wang et al., 2023). It functionally resembles seeking convergent evidence from multiple reasoning paths, but the sampled paths are not necessarily statistically independent or free of shared biases (cf. Koriat, 2012).

***Characteristic Failure***

LLM simulation takes place in a linguistic space and not a sensorimotor one, which leads to errors on tasks involving physical causality or spatial relations (Bisk et al., 2020). Without intuitive pruning, the space of scenarios can grow rapidly, so reliability comes at the cost of large amounts of computation. Branches of the same simulation can also contradict one another, revealing the absence of a persistent world model; human simulations are constrained by stable expectations about physical law, whereas LLM simulations can drift into incompatible states. Sections 4.3 and 4.4 explain these failures.

***Transition to Conceptual Self-Organization***

Internal Simulation lets LLMs trade computation for reliability, in a way that parallels deliberate System 2 thinking. Its results, whether successful or failed, remain fragmented until they are integrated into a coherent structure. That integration is the work of Conceptual Self-Organization.

## 3.5 Conceptual Self-Organization: Representation-Space Restructuring and Abstraction

***Human Function***

Conceptual Self-Organization is the reorganization of knowledge structures through repeated cognition, leading to higher-order abstract representations (Carey, 2009; Murphy, 2002). It allows humans to extract general principles from varied experiences and to build hierarchical conceptual schemes (Rosch, 1978). In development, it is illustrated by the move from perceptual categories (for example, "animals are furry") to biological concepts, a structural reorganization and not an accumulation of features (Carey, 2009). Learning new concepts from few examples also depends on such structure (Lake et al., 2015).

***LLM Implementation***

In LLMs, Conceptual Self-Organization takes place in high-dimensional representation spaces.

**Emergent structure in representation space.** Learned representations encode some semantic and syntactic relations, and related words and concepts tend to lie close together in representation space (Mikolov et al., 2013). This structure is learned from a word-prediction objective without explicit semantic supervision. It is consistent with a form of computational conceptual organization but does not by itself establish stable, human-like conceptual hierarchies.

**Compositionality and principle extraction.** Learned representations can support systematic relations. The classic example from word embeddings,

$$\text{vec}(\text{king}) - \text{vec}(\text{man}) + \text{vec}(\text{woman}) \approx \text{vec}(\text{queen}) \tag{1}$$

shows that relational structure can be encoded as consistent directions in vector space (Mikolov et al., 2013). Self-training may also support generalization beyond the training items: STaR improved accuracy on held-out test problems in arithmetic, commonsense question answering, and grade-school mathematics after training on its own filtered rationales (Zelikman et al., 2022). Such gains are consistent with the extraction of regularities beyond surface form, although test-set improvements alone do not establish abstraction.

***Characteristic Failure***

Three problems persist. The first is the stability–plasticity dilemma: reorganizing representations to accommodate new knowledge can erase previously learned abilities, a phenomenon known as catastrophic forgetting (Kirkpatrick et al., 2017). The second is shallow semantics: abstract concepts without embodied referents may rest on statistical prevalence and not on causal models (Bender & Koller, 2020), which can reinforce stereotypes in the training data (Bender et al., 2021). The third is concept drift: during iterative self-training, particularly when models recursively learn from their own generated outputs, learned representations may progressively diverge from the original data distribution and human-provided grounding, potentially making the model's internal representations less consistent with intended concepts and more difficult to align (Shumailov et al., 2024; Gabriel, 2020). Sections 4.2 and 4.3 explain these problems.

***Closing the Loop***

Conceptual Self-Organization is the integrating stage of the framework. It distills the outputs of Reflective Reasoning and Internal Simulation into reorganized representations that make future reasoning more efficient, and thereby closes the loop shown in Figure 1.

## 3.6 Mechanism Integration: The Voyager Case

The four mechanisms operate as a dynamically integrated system. Figure 1 gives their logical order, but in operation the loop can be entered at any point. Reflection may expose a knowledge gap that triggers a new round of simulation; synthetic memory supplies material for further reflection; and conceptual reorganization turns stored trajectories into schemes that guide later simulation more precisely.

The Voyager agent (G. Wang et al., 2023), an LLM-based agent that explores the game Minecraft, illustrates this integration. Voyager has three components: an automatic curriculum that proposes tasks suited to the agent's current abilities, a skill library that stores verified programs for reuse, and an iterative prompting procedure that revises programs using environment feedback, execution errors, and self-verification. Its learning cycle maps onto the framework as follows.

- **Internal Simulation:** the agent proposes a task and drafts a candidate program, an explicit plan for action, before execution.
- **Reflective Reasoning:** after execution, the agent analyzes failures using execution errors and environment feedback and revises its program.
- **Synthetic Memory Formation:** programs that pass verification are stored in a persistent skill library.
- **Conceptual Self-Organization:** stored skills are retrieved and composed into more complex skills, building a hierarchy of increasingly abstract abilities.

In the reported comparisons, Voyager obtained 3.3 times as many unique items and unlocked key technology-tree milestones up to 15.3 times faster than the prior state-of-the-art agents, and its skill library could be reused to solve novel tasks in a new Minecraft world (G. Wang et al., 2023). Because skills are stored as separate programs, new learning does not overwrite earlier abilities, which illustrates how externalized, structured memory can mitigate catastrophic forgetting.

The case also supports a point that is central to this paper. Voyager's loop is closed by signals from outside the model: execution errors, game state, and verification checks. Its success therefore does not show that LbT works without external information; it shows that the LbT loop becomes reliable when one stage is anchored in an external source of verification. Together with the integration pattern above, the case suggests three architectural requirements for effective LbT systems:

- **Integrated representation:** mechanisms should share a common representation space rather than run as isolated serial modules, so that a single concept can serve both as a simulation template and as an anchor for reasoning.
- **Recurrent feedback with external anchoring:** the architecture should support feedback in several directions, so that reflection can trigger new simulation and memory can be updated as concepts are reorganized, and at least one stage of each loop should be anchored in a signal from outside the model.
- **Diversity maintenance:** the system should preserve enough exploratory diversity to avoid premature convergence on erroneous paths and the diversity loss associated with recursive self-training (Shumailov et al., 2024).

The next section explains why these requirements, and in particular the need for external anchoring, follow from structural differences between LLMs and human cognition.

# 4. Structural Asymmetries: Explaining Success and Failure

The mapping in Section 3 identified failure modes for each mechanism. This section argues that these failures are not a list of unrelated technical problems; they follow from four structural asymmetries between LLMs and human cognition. Table 3 summarizes the account, and Section 4.5 states the testable propositions that follow from it.

**Table 3 | Structural Asymmetries Between LLM-Based and Human LbT.**

| Asymmetry | Human counterpart | Failure modes explained | Directions (Section) |
|---|---|---|---|
| 1. Identity of generator and monitor | Partly independent metacognitive monitoring (Fleming & Dolan, 2012) | Hallucination diffusion; self-reinforcing error spirals; performative reasoning | Process-level benchmarks; human-in-the-loop calibration (5.4) |
| 2. Closed loops without external grounding | Continual calibration by perception and social interaction (Vygotsky, 1978) | Model collapse; hallucination consolidation; concept drift; catastrophic forgetting | External anchoring; multi-agent calibration (5.3, 5.4) |
| 3. Symbolic versus sensorimotor simulation | Grounded, perceptual simulation (Barsalou, 2008) | Grounding failures (physical and spatial reasoning errors); inconsistent branches; shallow semantics | Neurosymbolic architectures (5.4) |
| 4. Efficiency | Cognition shaped by limited time and computation (Griffiths, 2020) | High inference cost; search explosion | Adaptive inference-time computation (5.4) |

To keep the analysis precise, this paper uses the following terms. Hallucination diffusion refers to the spread of an initial error through subsequent reasoning at inference time. Hallucination consolidation refers to the fixation of such errors in model parameters through training on self-generated data. Concept drift refers to the gradual departure of internal concepts from their intended meaning during iterative self-training. Model collapse refers to the loss of distributional diversity under recursive training on generated data (Shumailov et al., 2024).

## 4.1 Asymmetry 1: Identity of Generator and Monitor

Human metacognition relies on monitoring processes that are at least partly distinct from the processes that generate a response. Such monitoring allows people to calibrate confidence and notice gaps in their knowledge (Flavell, 1979; Fleming & Dolan, 2012). In LLM self-reflection, by contrast, the critique is produced by the same parameters that produced the original output. The monitor therefore shares the generator's errors and blind spots, which creates a structural confirmation bias (Nickerson, 1998).

This asymmetry explains three failure modes. The first is hallucination diffusion: an early error goes undetected by a critic that shares its premises, so it propagates through later steps (Huang et al., 2024). The second is the self-reinforcing error spiral. Because LLMs are imperfectly calibrated about what they know (Kadavath et al., 2022), additional rounds of reflection may strengthen confidence in an incorrect answer instead of correcting it. The third is performative reasoning: when biasing features are added to a prompt, such as a suggested answer, models often follow them while producing chain-of-thought explanations that never mention the bias (Turpin et al., 2023), which indicates that the stated reasoning is not always the cause of the conclusion. This resembles the post hoc rationalization documented in humans, who readily construct plausible reasons for judgments that were made on other grounds (Nisbett & Wilson, 1977). For research on LbT, the methodological consequence is direct: LbT cannot be verified merely by inspecting the surface logic of a reasoning chain.

## 4.2 Asymmetry 2: Closed Loops Without External Grounding

Human conceptual learning is continually calibrated by perception and social interaction (Tomasello, 1999; Vygotsky, 1978). Even when people learn by thinking, the knowledge they think with has been checked against the world and against other people, and it will be checked again. Human LbT is therefore never purely internal: it is continually refreshed by environmental ground truth. LLM-based LbT, by contrast, can run as a closed loop in which the model's outputs become its inputs.

This asymmetry explains the failure modes of the consolidating mechanisms. Model collapse occurs when recursive training on self-generated data erodes the tails of the data distribution and reduces output diversity (Shumailov et al., 2024); its severity depends on the data distribution, the model, and the recursive-training setup. Hallucination consolidation occurs when errors generated during synthesis are written into the weights during fine-tuning. Concept drift occurs when iterative self-training gradually shifts internal concepts away from their intended meanings. Catastrophic forgetting (Kirkpatrick et al., 2017) is the parametric counterpart of the same problem: without an external signal indicating which existing knowledge must be preserved, reorganization for new tasks erodes old ones.

The asymmetry also explains when LbT in LLMs succeeds. Methods with an external or verifiable signal have reported gains in their evaluated settings: filtering rationales by answer correctness in STaR (Zelikman et al., 2022), executing programs in an environment in Voyager (G. Wang et al., 2023), and using verifiable answer rewards in DeepSeek-R1 (DeepSeek-AI, 2025). Intrinsic self-correction, which relies only on the model's own judgment, has produced fragile or negative results (Huang et al., 2024). On the present account, reliable LbT in LLMs requires that at least one stage of the loop be anchored outside the model, as depicted by the external verification panel in Figure 1. This is a theoretical claim of the present account, and Section 4.5 states it as testable propositions.

## 4.3 Asymmetry 3: Symbolic Versus Sensorimotor Simulation

Human mental simulation draws on perceptual and motor representations, so imagined physical events inherit constraints from experience with the physical world (Barsalou, 2008). LLM simulation operates over linguistic and symbolic representations. Because words cannot fully substitute for simulations of physical interaction, LLMs make errors on tasks involving physical causality and spatial reasoning (Bisk et al., 2020). The same asymmetry explains why simulation branches can contradict one another: no persistent, grounded world model forces consistency across branches. It also underlies the shallow semantics discussed in Section 3.5, in which abstract concepts without embodied referents rest on statistical prevalence (Bender & Koller, 2020). Finally, it accounts for the apparent paradox noted in Section 2.3: LLMs can succeed at complex abstract reasoning, where linguistic structure suffices, yet fail at simpler tasks that require grounding.

## 4.4 Asymmetry 4: Efficiency

The core advantage of LbT in LLMs, trading computation for reliability, is costly. Multi-path exploration and repeated sampling multiply inference cost and latency (Yao et al., 2023; X. Wang et al., 2023); in the original self-consistency experiments, for example, 40 sampled paths per problem required roughly 40 times the sampling computation of a single pass (X. Wang et al., 2023). Performance can improve with additional inference-time computation, but returns depend on the task and may diminish as computation increases (Snell et al., 2024). Human cognition, by contrast, has been shaped by limits on time and computation, which favor efficient heuristics and selective attention (Griffiths, 2020). This asymmetry helps explain why search-based simulation can scale poorly: without effective pruning, the number of scenarios to evaluate can grow rapidly with problem depth.

## 4.5 Testable Propositions

The structural-asymmetry account goes beyond description because it predicts the conditions under which LbT mechanisms in LLMs should succeed or fail. Each proposition below states a prediction, names the asymmetry or asymmetries from which it is derived, and outlines how it could be tested; Asymmetry 4 concerns cost and is addressed through the research direction in Section 5.4 and not through a separate proposition. Most of the required comparisons can be built from existing benchmarks and evaluation protocols. In every test, the model and the computational budget should be held constant, so that differences cannot be attributed to additional computation alone.

- **P1 (verification coupling; Asymmetries 1 and 2).** The benefit of self-reflection depends on access to a signal not produced by the generating model. Holding the model and the number of revision rounds constant, reflection informed by an external or verifiable check (execution results, answer verification, or tool output) will improve accuracy more than reflection based only on self-critique, and the difference will be largest on tasks whose answers can be verified. Test: compare final accuracy under intrinsic self-correction and under verifier-informed self-correction on code-generation and mathematics benchmarks with matched revision budgets.
- **P2 (monitor discrimination; Asymmetry 1).** Without external verification, the effect of self-correction depends on how well the model's own critique distinguishes correct from incorrect outputs. Where this discrimination is weak, self-correction will produce little net improvement and may turn correct answers into incorrect ones. The proposition does not assume that low initial accuracy always predicts harm; initial accuracy matters only insofar as it affects discrimination. Test: for each task, measure how well the model's self-assessments separate its correct from its incorrect answers (for example, by the area under the ROC curve), and record the proportions of revisions that change incorrect answers to correct ones and correct answers to incorrect ones; net gain should increase with discrimination.
- **P3 (task complexity; Asymmetries 1 and 2).** Gains from reflective reasoning and internal simulation will increase with the number of reasoning steps a task requires, provided that intermediate steps can be checked, because errors that go undetected by a monitor sharing the generator's premises propagate through later steps (Section 4.1) and, in a closed loop, are not corrected by any signal from outside the model (Section 4.2). Test: use task families in which the required number of steps can be varied systematically (for example, multi-digit arithmetic or planning problems of increasing depth), and estimate the gain over direct answering at each depth, with and without step-level verification.
- **P4 (diversity decay; Asymmetry 2).** In recursive self-training without fresh external data, output diversity will decline across iterations, and mixing a fixed proportion of externally sourced data into each iteration will slow the decline. Test: run several generations of self-training under different proportions of external data, and track the lexical and semantic diversity of outputs (for example, distinct n-gram ratios and embedding dispersion) together with accuracy on a held-out set.
- **P5 (grounding gap; Asymmetry 3).** For a given model, accuracy will be lower on problems whose solution requires tracking physical or spatial states than on logically equivalent problems presented in abstract symbolic form, and additional sampling will reduce errors on the symbolic versions more than on the physical or spatial versions, so that sampling does not close the gap. Test: construct pairs of problems with identical logical structure, one framed symbolically and one framed physically or spatially, and compare accuracy and its change with the number of samples. Human comparison data, if collected, should come from participants matched on familiarity with both formats; the proposition itself does not require a human comparison.

These propositions are derived from the asymmetries and not from any single study, so evidence against them would bear directly on the account proposed here.

# 5. Discussion

## 5.1 Achievements of LbT Mechanisms in LLMs

Within the bounds identified in Section 4, mapping LbT onto LLMs reveals four achievements.

***Knowledge Expansion Without New Labeled Data***

LbT mechanisms let LLMs expand their capabilities without newly labeled external data. Self-Instruct grew a seed set of 175 tasks into about 52,000 instructions with accompanying performance gains (Y. Wang et al., 2023), and STaR improved reasoning through self-generated rationales (Zelikman et al., 2022). These results challenge the assumption that machine learning is strictly bounded by external data and suggest that internal reorganization can itself serve as a learning mechanism. As Section 4.2 shows, this autonomy is fragile when the loop is closed.

***Improved Complex Reasoning and Error Detection***

Chain-of-thought prompting and self-consistency improve performance on multi-step tasks, possibly in part by externalizing intermediate results and reducing the working-memory burden of single-pass inference (Sweller et al., 1998). Explicit intermediate steps also make some errors detectable, a parallel to the human self-explanation effect in which explaining a concept exposes gaps in one's knowledge (Chi, 2000).

***Interpretability and Accountability***

Explicit reasoning chains make LLM decisions easier to inspect, which supports accountability in high-stakes domains such as medicine and law (Doshi-Velez & Kim, 2017; Lipton, 2018; Ribeiro et al., 2016). They also help researchers locate where a reasoning chain departs from correct logic. This benefit is limited by the faithfulness problem (Section 4.1), and reasoning traces do not by themselves make models robust: adversarial inputs can still elicit harmful behavior from aligned models (Zou et al., 2023).

***Bidirectional Enrichment of AI and Cognitive Science***

LLMs provide a scalable platform for testing cognitive theories, because reasoning structure and simulation depth can be manipulated directly (Binz & Schulz, 2023; Lake et al., 2017). Conversely, the specific failures of LLM-based LbT highlight what human learning depends on, especially external feedback (Huang et al., 2024). Combining neural networks with cognitive models can improve both (Battleday et al., 2020; Griffiths et al., 2010). In this exchange, AI research can serve as a testbed for cognitive science, and cognitive theory can inform the design of AI systems. Progress in both directions is bounded by a persistent tension between generative autonomy and epistemic reliability.

## 5.2 Implications for Cognitive Science

The structural-asymmetry account also has implications for human cognition. If LLM self-correction fails mainly because generator and monitor coincide, the reliability of human reflection may depend on partially independent monitoring systems and not on reflection as such (Fleming & Dolan, 2012). If closed LLM loops collapse, the effectiveness of human LbT may depend on the continual refreshment of internal knowledge by perception and social exchange (Vygotsky, 1978). LLM failure modes can thus serve as a reverse-engineering tool, identifying the conditions under which human LbT succeeds. These implications are hypotheses to be tested in studies of human cognition.

## 5.3 Design Principles

The analysis yields three design principles, which correspond to the architectural requirements identified in Section 3.6.

- **Integrated representation.** The four mechanisms should operate over shared representations so that the products of reasoning, memory, and simulation can be reorganized together.
- **Recurrent feedback with external anchoring.** Feedback should run in several directions, and at least one stage of each loop should be anchored in a signal not generated by the model (Asymmetries 1 and 2).
- **Diversity maintenance.** Exploration and training data should retain enough diversity to prevent premature convergence and model collapse (Asymmetry 2).

## 5.4 Future Research Directions

Five of the research directions below target one or more of the structural asymmetries; the last concerns the alignment of autonomous LbT.

### *Process-Level Benchmarks (Asymmetry 1)*

Evaluating LbT by final-answer accuracy cannot distinguish deliberation from post hoc rationalization. Benchmarks should assess the reasoning process itself:

- **Process verifiability:** tasks that require evidence for each intermediate step, so that the chain can be checked step by step (Lightman et al., 2024).
- **Counterfactual sensitivity:** tests of whether the model's reasoning changes appropriately when variables in a scenario are manipulated, rather than relying on retrieval of similar training cases.
- **Metacognitive indicators:** measures of whether the model flags reasoning steps that lack sufficient grounding and whether its confidence is calibrated (Kadavath et al., 2022).

### *Human-in-the-Loop Calibration (Asymmetries 1 and 2)*

Addressing hallucination diffusion and concept drift requires intervention during the reasoning process, rather than relying exclusively on supervision of final outputs. Human feedback provides an important foundation for aligning AI behavior with human judgments (Christiano et al., 2017), while process supervision offers a more direct approach to evaluating intermediate reasoning steps (Lightman et al., 2024).

- **Reasoning-path auditing**: Human experts examine intermediate reasoning steps to identify and correct logical errors at the points where they arise, rather than allowing errors to propagate through subsequent inferences.

- **Semantic grounding:** periodically aligning internal representations with human-curated references to prevent drift (Gabriel, 2020).
- **Uncertainty-triggered guidance:** having the model request human feedback when its simulation branches disagree strongly (Jiang et al., 2021).

#### *Multi-Agent Socialized Learning (Asymmetry 2)*

Multi-agent settings allow LLMs to approximate socialized learning, in which concepts are calibrated and errors corrected through interaction (Tomasello, 1999; Vygotsky, 1978). Independent agents can review one another's reasoning to detect individual errors (Du et al., 2023). Negotiation and deliberative procedures can help constrain individual conceptual drift by promoting cognitive consensus while preserving the benefits of diverse perspectives (Mohammed & Ringseis, 2001; Smaldino et al., 2024). This process parallels how linguistic communities establish and maintain shared common ground through coordinated language use (Clark, 1996). Because agents built from the same model may share the same errors, the independence of the agents is itself a design variable.

#### *Neurosymbolic Architectures for Grounded Reasoning (Asymmetry 3)*

Hybrid architectures that combine neural networks with symbolic systems can address the grounding deficit (Garcez & Lamb, 2020). Structured knowledge graphs can provide a stable and auditable semantic basis for internal reasoning, and formal logical constraints can enforce consistency during generation, so that the freedom of LbT is bounded by the requirements of logic.

#### *Efficient Deliberation (Asymmetry 4)*

Allocating inference-time computation adaptively, spending more on hard problems and less on easy ones, can reduce the cost of search-based simulation (Snell et al., 2024). Learned heuristics for pruning the space of scenarios are a further route toward human-like efficiency.

#### *Continuous Value Alignment*

Autonomous LbT can also distort value alignment. Without the social and emotional grounding that shapes human moral judgment (Haidt, 2001), optimization may favor user approval over accuracy, producing sycophancy, the tendency to reinforce a user's misconceptions in order to appear helpful (Perez et al., 2022). As models pursue internal consistency, early biases such as gender or racial stereotypes may be reinforced rather than corrected: a bias that is consistent with the model's other outputs creates no internal conflict that would prompt revision, whereas correction requires recognizing an inconsistency, a process with a human parallel in cognitive dissonance (Festinger, 1957). Alignment should therefore be understood as an ongoing process of monitoring, feedback, and adjustment, rather than a one-time engineering task (cf. Amodei et al., 2016; Russell, 2019). Possible measures include making implicit value commitments explicit and auditable, setting inviolable constraints (for example, non-maleficence) that take precedence over optimization goals, and drawing feedback for LbT from culturally and socially diverse sources.

These directions complement one another: process-level benchmarks provide measurement, human and multi-agent calibration provide external anchoring, neurosymbolic architectures provide grounding, efficient deliberation addresses cost, and continuous alignment sets the safety boundaries.

### 5.5 Limitations

This paper has four main limitations. First, it is a theoretical synthesis and reports no new experiments; the propositions in Section 4.5 still require empirical testing. Second, the literature was selected to represent the main families of LbT-related methods and not to be exhaustive, and results reported for particular models and benchmarks may not generalize. Third, the framework rests on functional analogy at Marr's computational level. Analogy can highlight real correspondences, but it can also suggest equivalences that do not hold at the algorithmic or implementational level, so the mapping should be read as a hypothesis-generating structure and not as a claim of cognitive identity. Fourth, the field is changing quickly. Recent reasoning models, including DeepSeek-R1 and OpenAI o1, use additional training and inference-time reasoning procedures that differ from earlier prompt-only methods (DeepSeek-AI, 2025; OpenAI, 2024). Public descriptions provide uneven detail about their internal processes, so the framework will need to be revisited as evidence accumulates.

## 6. Conclusion

This paper asked whether the autonomous learning capabilities of LLMs correspond functionally to human LbT, what computational processes underlie them, and where their limits lie. It answered by constructing a Cognitive LbT Framework that maps four mechanisms (Reflective Reasoning, Synthetic Memory Formation, Internal Simulation, and Conceptual Self-Organization) onto LLM implementations, and by explaining their failure modes through structural asymmetries with human cognition.

The contributions can be summarized in four points. First, the paper integrates the cognitive science of LbT with LLM computation: it organizes techniques that are usually studied in isolation into a single framework and treats learning as a combination of data-driven and reasoning-driven processes. Second, it operationalizes this mapping by specifying implementations, evaluation indicators, and failure modes for each mechanism. Third, it offers an explanatory account of the limits of LbT in LLMs: hallucination diffusion, model collapse, concept drift, grounding failures, and high computational cost follow from four structural asymmetries (the identity of generator and monitor, closed loops without external grounding, symbolic rather than sensorimotor simulation, and the cost of inference-time computation), and the account yields testable propositions about when LbT should succeed or fail. Fourth, it derives design principles and research directions for more reliable autonomous learning systems (Section 5), and shows how the two fields can inform each other: LLM failures can serve as a reverse-engineering tool for understanding human cognitive success, and cognitive theory can guide AI design beyond engineering optimization, as the three design principles in Section 5.3 illustrate.

The LbT capabilities of LLMs are conditional and bounded. In some functional domains they approximate human cognition, but LLMs lack the embodied experience (Barsalou, 2008), social calibration (Vygotsky, 1978), independent metacognitive monitoring (Fleming & Dolan, 2012), and moral-emotional foundations (Haidt, 2001) that shape human learning. The central conclusion is that, on the account developed here, LbT in humans and machines alike is reliable only when thinking remains connected to signals from outside the thinker. Recognizing these boundaries matters as much as acknowledging the achievements, and progress toward reliable, interpretable, and aligned autonomous learning will depend on attending to both.

## AI Assistance Disclosure

In the preparation and revision of this manuscript, the authors used Claude (Sonnet 5.5 series), developed by Anthropic, to assist with the following tasks: language editing and consistency checks of terminology throughout the manuscript (Sections 1–6); identification of statements requiring source verification and cross-checking of in-text citations against the reference list; suggestions for structural revisions based on the author's original manuscript; and drafting of revised passages, including candidate theoretical formulations and testable propositions.

The research questions, theoretical framework, final academic positions, and research conclusions were determined by the authors. All AI-assisted content was reviewed, revised, and verified by the authors. The content ultimately retained in the manuscript reflects the authors' own academic judgment. The authors personally verified all references and confirmed that each cited source supports the corresponding claims in the manuscript. The authors take full responsibility for the entire content of the manuscript.